\documentclass[runningheads]{llncs}

\usepackage{eccv}

\usepackage{eccvabbrv}

\usepackage{graphicx}
\usepackage{booktabs}

\usepackage[accsupp]{axessibility}  

\usepackage{hyperref}

\begin{document}

\title{Hierarchical and Holistic Open-Vocabulary Functional 3D Scene Graphs for Indoor Spaces}

\titlerunning{Open-Vocabulary Functional 3D Scene Graphs}

\author{
    Xinggang Hu$^{1,2\star}$ \and
    Chenyangguang Zhang$^{3\star}$ \and
    Alexandros Delitzas$^{3,4}$ \and
    \mbox{Xiangkui Zhang$^{2}$} \and
    Marc Pollefeys$^{3,5}$ \and
    Francis Engelmann$^{6}$ \and
    Xiangyang Ji$^{1\dagger}$
}

\authorrunning{X.~Hu et al.}

\institute{
    $^1$Tsinghua University, China \quad
    $^2$Dalian University of Technology, China \\
    $^3$ETH Zurich, Switzerland \quad
    $^4$MPI for Informatics, Germany \\
    $^5$Microsoft, Switzerland \quad
    $^6$USI Lugano, Switzerland
}

\maketitle

\begingroup
\let\thefootnote\relax
\footnotetext{\raggedright \textsuperscript{*}Equal contribution. \textsuperscript{\ensuremath{\dagger}}Corresponding author.}
\endgroup

\begin{abstract}
Functional 3D scene graphs offer a versatile and flexible representation for 3D scene understanding and robotic manipulation, defined by object nodes, interactive elements, and functional relationship edges. 
However, their potential remains underexplored due to the limited coverage of existing benchmarks and the overly straightforward design of previous pipelines, which primarily focus on large-scale furniture but lack of hierarchical structures. 
Therefore, in this work, we extend the benchmark coverage by introducing dense tabletop objects and explicit multi-level functional relationships. 
This expansion introduces critical challenges involving small-scale, dense, and similar instances, with lack of visual anchoring in relational reasoning, instance confusion during cross-frame fusion, and attribution uncertainty under dynamic viewpoints. 
To address these issues, we propose an open-vocabulary pipeline based on 2D visual grounding and 3D graph optimization. 
Specifically, we anchor fine-grained functional edges from 2D visual evidence, and associate nodes across frames in 3D using multiple cues. 
Furthermore, edge association is formulated as temporal graph optimization, integrating evidence accumulation, entropy regularization, and temporal smoothing to robustly determine the functional connections of each node. 
Finally, global hierarchy shaping is performed to recover the hierarchical graph structure.
Extensive experiments demonstrate that the proposed method can reliably infer functional 3D scene graphs in challenging real-world scenes, thereby further unlocking their potential for practical applications.
Code is available at \url{https://github.com/Hbelief1998/HHOpenFunGraph-ECCV26}.
  \keywords{3D Scene Graph \and Scene Understanding \and Open Vocabulary}
\end{abstract}

\section{Introduction}

As virtual reality, robotics, and embodied intelligence continue to advance, it becomes increasingly vital to develop a versatile and accurate 3D scene representation that bridges the gap between high-level 3D understanding and downstream tasks such as manipulation and content generation.
Traditionally, 3D scene graphs \cite{chen2024clip,koch2024open3dsg,gu2024conceptgraphs,armeni20193d,koch2024lang3dsg,wu2021scenegraphfusion,
rosinol20203d,rosinol2021kimera,wald2020learning} have been proposed as discrete graph-based representations of 3D environments, consisting of object nodes and their spatial relationships. 
While such representations have proven effective for robotic navigation and scene understanding, their coarse granularity limits their applicability to fine-grained tasks such as manipulation.

\begin{figure}[t]
  \centering
   \includegraphics[width=1.0\linewidth]{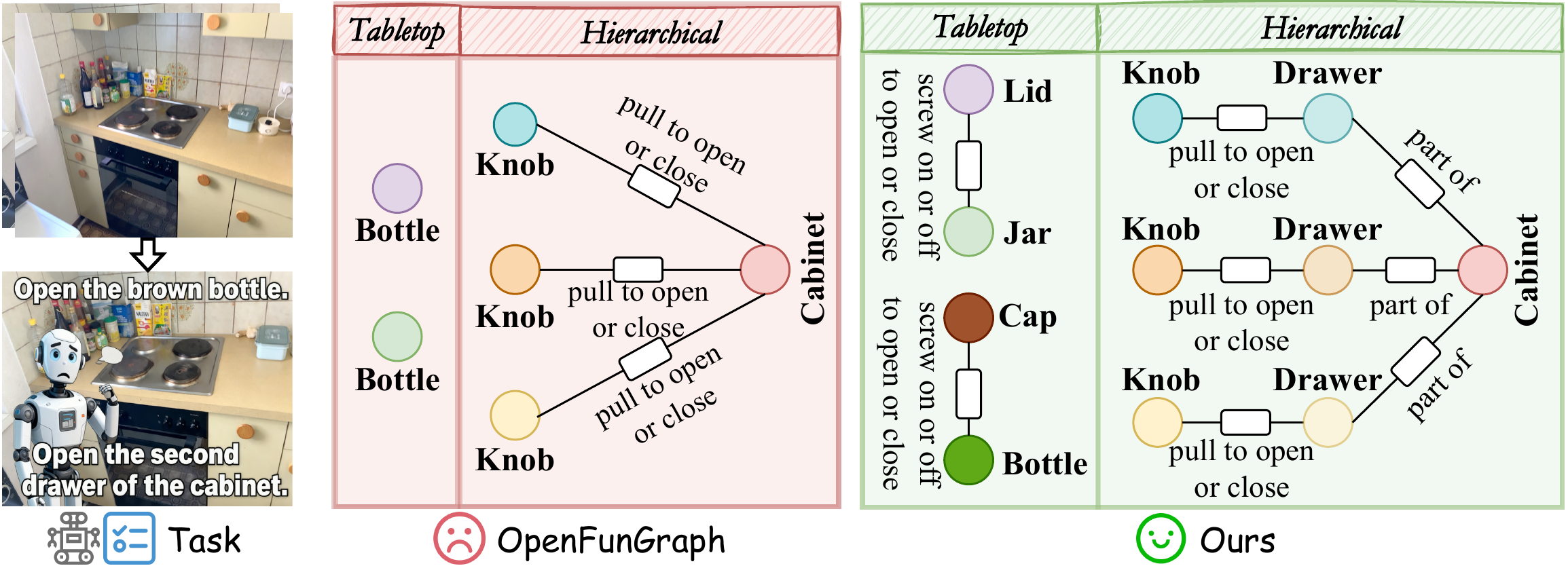}
   \caption{Hierarchical and holistic functional 3D scene graphs. 
   In contrast to prior approaches~\cite{zhang2025open}, we model tabletop manipulable objects and explicit hierarchical object–part structures in functional 3D scene graphs.
   }
   \label{fig:teaser}
\end{figure}

Recently, OpenFunGraph~\cite{zhang2025open} introduced the concept of functional 3D scene graphs by extending traditional scene graphs to include objects, interactive elements, and functional relationships.
Although this representation is theoretically flexible and capable of encoding rich semantic and relational properties, its benchmark and pipeline remain limited.
The existing setup cannot handle the complexity commonly observed in real-world scenarios.
Specifically, the annotations in SceneFun3D \cite{delitzas2024scenefun3d} and FunGraph3D \cite{zhang2025open} consider only large furniture items (\textit{e.g.}, cabinets, bathtubs, wardrobes) and lack the tabletop objects that are more common in robotic manipulation tasks. 
In addition, the intrinsic partonomy and hierarchical structure of objects and parts are not modeled, restricting the applicability of the representation to fine-grained reasoning. 
For example, as Figure~\ref{fig:teaser} shows, multiple knobs are ambiguously linked to the same cabinet without distinction between drawers, leading to confusion in manipulation queries (\textit{e.g.}, open the second drawer of the cabinet).

To overcome these limitations, we extend the capabilities of the functional 3D scene graph benchmark by introducing: 
(1) diverse tabletop objects and their associated functional attributes (\textit{e.g.}, bottle-cap, kettle-handle), covering small, manipulable targets commonly found in real-world operations; 
(2) explicit hierarchical relationships between objects and their constituent parts, treating intermediate structures as functional carriers (\textit{e.g.}, control panel, drawer, lid), and end-effector components that can be directly manipulated as interactive units (\textit{e.g.}, switch, handle, knob), thereby characterizing typical multi-level structures such as oven-panel-switch, cabinet-drawer-handle, and pot-lid-knob.

In real-world scenarios, introducing tabletop objects and hierarchical structures is crucial for downstream applications. However, this benchmark expansion introduces perception and reasoning challenges that the existing paradigm of OpenFunGraph~\cite{zhang2025open} struggles to address:
(1) {Lack of Visual Anchoring in Relational Reasoning.} The 3D geometry of tiny tabletop parts (\textit{e.g.}, bottle caps and kettle handles) is highly susceptible to reconstruction noise, while hierarchical structures introduce a massive number of fine-grained nodes of the same category in close spatial proximity (\textit{e.g.}, drawers), resulting in severe 3D bounding box aliasing. Despite such geometric degradation, OpenFunGraph still feeds these distorted 3D bounding boxes in natural language format for an LLM to infer relationships~\cite{zhang2025open}. The compounding of unreliable 3D information with the inherent deficiency of the LLM in reasoning over numerical spaces leads to severe error accumulation, rendering it difficult to robustly establish accurate functional connections in the absence of 2D visual evidence.
(2) {Instance Confusion in Cross-Frame Fusion.} In tabletop and hierarchical scenes, dense instances of the same category are prone to boundary and feature overlap during cross-frame fusion (\textit{e.g.}, compactly arranged spice bottles in a kitchen). OpenFunGraph primarily relies on geometric proximity and performs average aggregation of CLIP similarities~\cite{zhang2025open}, which easily leads to instance aliasing.
(3) {Attribution Uncertainty.} Tabletop scenes feature dense arrangements of similar instances with mutual occlusions, while fine-grained nodes introduced by hierarchical structures often lack sufficient distinctiveness from certain viewpoints (\textit{e.g.}, blurred boundaries between cabinet drawers and doors). These issues cause the attribution of interaction units to be highly uncertain among similar neighbors, frequently flipping with viewpoint changes. Relying solely on instantaneous scoring like OpenFunGraph makes edge association prone to jitter and erroneous connections.

To overcome the challenges introduced by the tabletop and hierarchical settings in the expanded benchmark, we propose an open-vocabulary pipeline based on direct visual grounding and graph optimization for constructing hierarchical and holistic functional 3D scene graphs, as illustrated in Figure~\ref{fig:teaser}.  
First, to address the {lack of visual anchoring in relational reasoning}, guided by an interactability map constructed via LLMs and open-vocabulary detection, we utilize visual foundation models to generate candidates for multi-level nodes. By fusing 2D geometry with visual-semantic consistency, we infer 2D functional edges from object nodes to fine-grained nodes based on direct 2D visual evidence.
Second, to mitigate {instance confusion during cross-frame fusion}, we introduce a multi-cue node association mechanism. This integrates appearance, semantics, 2D projection, and 3D geometric information to suppress erroneous associations among dense instances of the same category, thereby forming stable cross-frame node trajectories.
Third, regarding {attribution uncertainty under dynamic viewpoints}, we model edge association as time-evolving variables within a graph optimization framework. Incorporating historical evidence accumulation, entropy regularization, and temporal smoothing, this optimization robustly determines edge connections amidst candidate competition caused by viewpoint changes and observation noise, effectively preventing association jitter and erroneous connections.
Finally, a hierarchy completion step refines the structure by optimizing semantic and geometric consistency, resulting in the final hierarchical graph.

Experiments on our expanded benchmark demonstrate our method's effectiveness in complex scenes featuring tabletop objects and hierarchical structures, outperforming the FunGraph~\cite{rotondi2025fungraph} and modified OpenFunGraph~\cite{zhang2025open} baselines by up to 13.9\% and 20.5\% in node and edge triplet retrieval, respectively.

\section{Related Works}
\noindent\textbf{3D Scene Understanding.}
3D indoor scene understanding has long focused on closed-set 3D semantic/instance segmentation~\cite{atzmon2018point,choy20194d,hu2021vmnet,hua2018pointwise,huang2023segment3d,landrieu2018large,weder2023alster,li2018pointcnn,qi2017pointnet,qi2017pointnet++,thomas2019kpconv,weder2024labelmaker,engelmann20203d,han2020occuseg,hou20193d,takmaz20233d,jiang2020pointgroup,schult2023mask3d,vu2022softgroup,yue2023agile3d}. Subsequent work, driven by vision-language foundation models, has extended to open-vocabulary 3D semantic segmentation~\cite{engelmann2024opennerf,kerr2023lerf,jatavallabhula2023conceptfusion,peng2023openscene,takmaz2023openmask3d,zhou2024feature,zuo2024fmgs,yilmaz2024opendas,qin2024langsplat,takmaz2025search3d} and 3D vision-language alignment~\cite{huang2022multi,yang2021sat,hsu2023ns3d,opencity3d2025,zhang2023multi3drefer,roh2022languagerefer,Parelli_2023_CVPR,Delitzas_2023_BMVC}. 
Recent part-level open-vocabulary methods further recognize fine-grained object parts from open textual descriptions~\cite{sun2023going,wei2023ov}, highlighting the need to move beyond whole-object perception. However, they mainly focus on part segmentation or recognition, rather than temporally fused 3D functional entities and graph-structured functional relationships. Unlike these perception-oriented works, SceneFun3D~\cite{delitzas2024scenefun3d} proposed a benchmark containing interactive elements. Building on this, FunGraph3D~\cite{zhang2025open} further provides object annotations and the relationships between interactive elements and objects.
However, this benchmark does not yet consider tabletop manipulable objects or the hierarchical structure between objects and parts. This paper extends this benchmark by introducing more holistic tabletop objects and hierarchical annotations.

\noindent\textbf{Affordance Understanding.}
Affordance understanding concerns how an environment can be operated upon. Existing methods predominantly predict affordance heatmaps from images~\cite{do2018affordancenet,zhai2022one}, videos~\cite{fang2018demo2vec,nagarajan2019grounded,yoshida2024text}, or 3D representations~\cite{cho2024dense,banerjee2024introducing,ye2024g,fan2024hold,zhang2024moho,zhang2024ddf,ye2022s,chen2022alignsdf}, or learn interaction strategies from human-scene interaction demonstrations~\cite{cho2024dense,banerjee2024introducing,ye2024g,fan2024hold,zhang2024moho,zhang2024ddf,delitzas2026funrec,ye2022s,chen2022alignsdf}. 
However, existing approaches typically attach affordances to the object holistically, merely addressing ``how a certain object can be used,'' failing to specify the concrete operational parts.
In contrast, we explicitly bind affordances to specific interactive parts, uniformly modeling hierarchical operational relationships via functional edges.

\noindent\textbf{3D Scene Graphs.}
3D scene graphs organize indoor entities and their spatial/semantic relationships in a node-edge format~\cite{armeni20193d,rosinol20203d,rosinol2021kimera,wald2020learning,koch2024lang3dsg,wang2023vl,wu2021scenegraphfusion,wu2023incremental,zhang2021exploiting,zhang2021knowledge,takmaz2025search3d}. Building on this, functional 3D scene graphs further incorporate interactive elements as nodes, modeling the functional connections between objects and interactive elements. 
IFR-Explore~\cite{li2021ifr} mines functional relationships in synthetic scenes using reinforcement learning, but is limited by a closed-set setting and ignores part-level elements; open-vocabulary 3D scene graph methods such as Open3DSG~\cite{koch2024open3dsg} and ConceptGraph~\cite{gu2024conceptgraphs} primarily remain at the object-level, addressing only object nodes and a few spatial relationships.
Furthermore, FunGraph~\cite{rotondi2025fungraph} introduces functional part nodes and \texttt{has-part} edges into the 3D scene graph through 2D functional part detection and VLM annotation,
OpenFunGraph~\cite{zhang2025open} builds upon this by employing open-vocabulary object and interactive element detection and sequentially inferring functional relationships. However, both have limited coverage of tabletop manipulable objects and do not explicitly model the hierarchical object–part structure.
To address the aforementioned limitations, this paper constructs a functional 3D scene graph that simultaneously possesses a multi-level structure and broader coverage of manipulable objects, through 2D functional relationship anchoring, cross-frame node association, functional graph optimization, and global graph refinement.

\section{Problem Definition}

This work aims to address the task of automatically constructing a hierarchical functional 3D scene graph $\mathcal{G}$ for holistic items in general indoor scenes given a sequence of posed RGB-D images. 

The graph $\mathcal{G}=(\mathcal{V},\mathcal{E})$ is defined with: 
(1) The node set $\mathcal{V}=\mathcal{O}\cup\mathcal{C}\cup\mathcal{U}$, where Objects $\mathcal{O}$ are the ultimate hosts of functionality and serve as interaction targets (\textit{e.g.}, cabinet, pot, oven, etc.); 
functional carriers $\mathcal{C}$ are structures on objects that undertake intermediate functions or operable mid-level units (\textit{e.g.}, drawer, lid, control panel, etc.); 
interactive units $\mathcal{U}$ are the end-effector elements that can be directly manipulated by a user or robot to trigger functions (\textit{e.g.}, handle, knob, switch, etc.).
(2) The directed edge set $\mathcal{E}$ represents functional relationships, characterizing the compositional, subordinate, or control interactions among nodes. These relationships constitute a functional hierarchy: depending on the observability and modeling of the functional carrier layer $\mathcal{C}$, two functional relational structures emerge, namely $\mathcal{O} \leftarrow \mathcal{C} \leftarrow \mathcal{U}$ (\textit{e.g.}, oven $\leftarrow$ control panel $\leftarrow$ knob) and $\mathcal{O} \leftarrow \mathcal{U}$ (\textit{e.g.}, bottle $\leftarrow$ cap). 
The objective of this work is to jointly infer the 3D properties (\textit{e.g.}, position, dimension, and semantics) of objects $\mathcal{O}$, functional carriers $\mathcal{C}$, and interaction trigger units $\mathcal{U}$, alongside their functional relationships $\mathcal{E}$, to ultimately construct a complete scene graph $\mathcal{G}$.

\section{Method}

\begin{figure*}[t]
  \centering
   \includegraphics[width=1.0\linewidth]{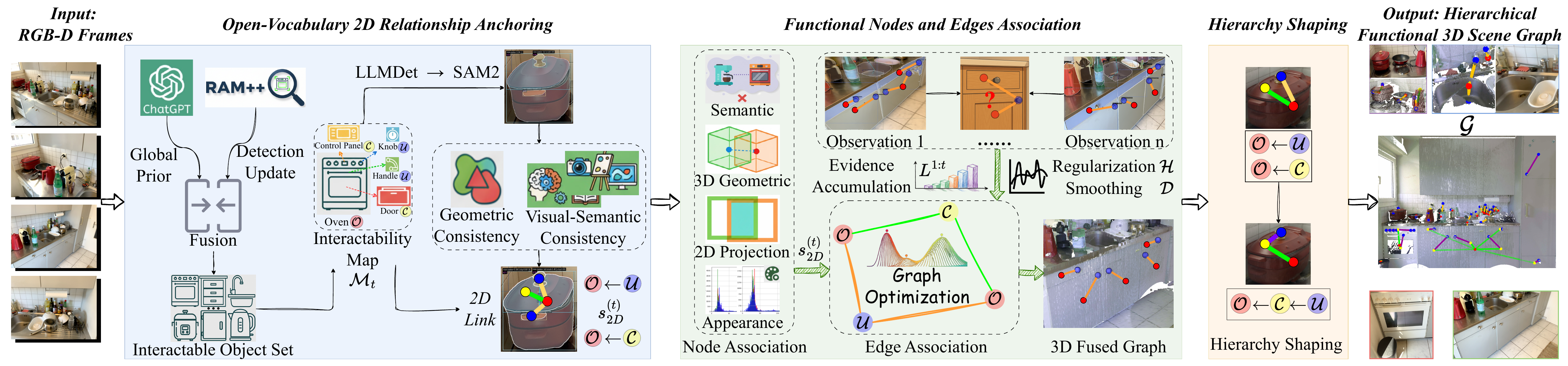}
   \caption{Overview of the hierarchical functional 3D scene graph construction pipeline. 
   }
   \label{fig:Pipeline}
\end{figure*}

To jointly infer the 3D attributes (position, size, semantics, etc.) of objects $\mathcal{O}$, functional carriers $\mathcal{C}$, and interactive units $\mathcal{U}$, as well as the relationships $\mathcal{E}$ between them, thereby constructing the holistic hierarchical functional 3D scene graph $\mathcal{G}$, this study proposes a three-stage progressive reasoning framework, shown in Figure~\ref{fig:Pipeline}.
First, in the 2D relation anchoring stage, we construct a dynamic interactability map to guide the open-vocabulary instance detection and segmentation of nodes $\mathcal{V}=\mathcal{O}\cup\mathcal{C}\cup\mathcal{U}$, and generate 2D functional edges based on visual evidence by integrating geometric and visual-semantic consistency evaluations.  
Next, we associate nodes across frames by fusing appearance, semantics, 2D projection consistency, and 3D geometric information to obtain temporally consistent global node trajectories. 
The corresponding candidate edges are modeled as time-evolving graph variables, and the functional edges $\mathcal{E}$ are robustly determined using a graph optimization strategy that combines cross-frame evidence accumulation with entropy regularization and temporal smoothing.
Finally, in the global graph construction stage, we perform hierarchy shaping on the associated functional edges to explicitly model the functional hierarchy of $\mathcal{O} \leftarrow \mathcal{C} \leftarrow \mathcal{U}$, ultimately outputting the complete $\mathcal{G}$.

\subsection{Open-Vocabulary 2D Relationship Anchoring}
The compounding of 3D geometric degradation and the inherent spatial reasoning deficiencies of LLMs leads to severe error accumulation. 
Specifically, tabletop tiny parts suffer from unstable localization under reconstruction noise, and dense hierarchical nodes (\textit{e.g.}, adjacent drawers) exhibit severe bounding box aliasing. Consequently, 3D spatial proximity becomes non-discriminative.
To circumvent this issue, we abandon the paradigm of relying on distorted 3D coordinates for inference and instead directly extract visual evidence from 2D frames. This stage aims to generate reliable functional edge candidates and their associated visual evidence scores, providing an unambiguous basis for subsequent 3D graph fusion.

\noindent\textbf{Generation of Dynamic Interactability Map.}
To guide open-vocabulary instance detection and relationship reasoning, we introduce a Dynamic Interactability Map $\mathcal{M}_t$, which specifies the manipulable objects $\mathcal{O}$ identified in the current frame $t$ and their associated functional component ($\mathcal{C}, \mathcal{U}$) types.
The construction of $\mathcal{M}_t$ fuses the general knowledge of an LLM with the visual evidence from the current frame. 
First, we utilize the LLM GPT-5~\cite{openai_gpt5_system_card_2025} based on task definitions to generate a prior list of globally manipulable objects (\textit{e.g.}, cabinet, oven, washing machine, kettle, etc.). However, this global prior can hardly exhaust scene-specific fine-grained categories and does not reflect frame-level visibility. 
Therefore, we subsequently identify instances at frame $t$ using RAM++~\cite{zhang2024recognize, huang2025open}, filter for manipulability via an LLM, and integrate global priors to determine the set of manipulable objects $\mathcal{O}$ for the current frame.
Finally, we prompt the LLM, for the objects in $\mathcal{O}$, to generate their associated functional carrier $\mathcal{C}$ and interactive unit $\mathcal{U}$ types, thus updating $\mathcal{M}_t$ dynamically.

\noindent\textbf{Open-Vocabulary Node Candidate Generation.}
Guided by $\mathcal{M}_t$, we employ a two-stage pipeline to generate the set of detected nodes for the current frame, denoted as $\mathcal{N}_t=\{n_k\}$. LLMDet~\cite{fu2025llmdet} generates bounding boxes $\mathbf{b}_{n_k}$, categories $c_{n_k}$, and confidence scores $s_{n_k}$ conditioning on $\mathcal{M}_t$. Subsequently, we prompt SAM2~\cite{ravi2024sam2} with $\mathbf{b}_{n_k}$ to yield masks $M_{n_k}$ and extract appearance features $\mathbf{h}_{n_k}$. Each node $n_k$ comprises attributes including $(\mathbf{b}_{n_k}, c_{n_k}, s_{n_k}, M_{n_k}, \mathbf{h}_{n_k})$.

\noindent\textbf{2D Functional Edge Generation and Scoring.}\label{Meth:2DFEG}
Based on the detected node $\mathcal{N}_t$ and the Dynamic Interactability Map $\mathcal{M}_t$, this stage aims to construct and score functional edge $(o \leftarrow f)$ candidates, where $o \in \mathcal{O}$ is an object node and $f \in \mathcal{C}\cup\mathcal{U}$ is a fine-grained node.
Note that we do not directly construct the $\mathcal{O} \leftarrow \mathcal{C} \leftarrow \mathcal{U}$ hierarchy in the 2D stage due to 1) increased combinatorial complexity in candidate generation and 2) high ambiguity under single-frame observations. 
Instead, it is deferred to the global hierarchy shaping stage (Section~\ref{Meth:GHS}), which utilizes global 3D information after the sequence processing is complete.

The generation of candidate pairs $(o, f)$ should first follow the semantic constraints provided by $\mathcal{M}_t$. 
Subsequently, we employ a cascaded verification process: candidate pairs must first pass a pre-filtering based on node confidence ($\mathcal{S}_{\text{det}}=\min(\text{s}_o, \text{s}_f) > \tau_{\text{det}}$) and geometric consistency ($\mathcal{G}_{\text{camc}} > \tau_{\text{geo}}$). 
The thresholds are set to $\tau_{\text{det}}=0.25$ and $\tau_{\text{geo}}=0.90$ to cover as many detection candidates as possible, while completing the filtering through strict geometric constraints, thus balancing comprehensiveness and accuracy.
Here, $\mathcal{G}_{\mathrm{camc}} = \frac{\lvert M_f \cap \big(\hat M_o\oplus\Delta\big)\rvert}{\lvert M_f\rvert}$ ensures that the mask $M_f$ is contained within the mask $\hat{M}_o$ ($\oplus\Delta$ denotes a pixel-level morphological dilation on the object mask to tolerate contour errors).
Candidate pairs $(o \leftarrow f)$ that pass the pre-filtering are then subjected to a final visual-semantic consistency assessment by the VLM LLAVA v1.6~\cite{liu2024visual,liu2024improved,liu2024llava} to generate the 2D functional edge score $s_{2D}^{(t)}(o \leftarrow f) \in (0, 1]$. 
$s_{2D}^{(t)}$ quantifies the support of the current frame for this $(o \leftarrow f)$ connection and will serve as a key input for the 3D spatiotemporal evidence accumulation.

\subsection{Functional Nodes and Edges Association}\label{Met:FNEA}

Although reliable single-frame 2D evidence is extracted in the previous stage, lifting it to 3D introduces critical challenges arising from tabletop and hierarchical settings: cross-frame instance confusion and attribution uncertainty. The small-scale, densely arranged nodes with prone-to-overlap boundaries (\textit{e.g.}, bottles, drawers) inherent to tabletop and hierarchical structures often lead to erroneous cross-frame associations. Furthermore, single-frame functional edges are unstable due to occlusions and viewpoint changes, causing fine-grained nodes to exhibit attribution competition among multiple neighboring candidates (\textit{e.g.}, handle attribution at the junction of cabinet opening structures), making edge states highly susceptible to noise-induced jitter. To address this, we first robustly associate nodes via multi-cue fusion, and subsequently model functional edges as time-evolving graph variables. By jointly constraining long-term evidence accumulation, entropy regularization, and temporal smoothing, we robustly determine functional connections within a graph optimization framework.

\noindent\textbf{Node Association.}
In frame $t$, to associate the set of detected nodes with 2D functional edges $\mathcal{N}_t = \{n_k\}$ with the set of node instances in the map $\mathcal{V} = \{v_j\}$, we define a bipartite graph edge weight $S^{(t)}(n_k \to v_j)$:
\begin{equation}
\begin{split}
S^{(t)} = & \, w_{\text{iou}} \text{IoU}(\mathbf{b}_{n_k}, \Pi_t(\mathcal{P}_{v_j})) + w_{\text{geo}} \mathcal{K}(\|\mathbf{x}_{n_k}^{3D} - \boldsymbol{\mu}_{v_j}\|) \\
          & + w_{\text{app}} \cos(\mathbf{h}_{n_k}, \mathbf{H}_{v_j}) + w_{\text{sem}} \mathcal{J}(c_{n_k}, c_{v_j}).
\end{split}
\end{equation}
Here, $\mathcal{P}_{v_j}$ denotes the accumulated 3D point cloud of the map node $v_j$, $\Pi_t(\mathcal{P}_{v_j})$ denotes its projected 2D bounding box in frame $t$, and $\mathbf{H}_{v_j}$ denotes its accumulated appearance histogram updated over time.
We set $w_{\text{iou}}=0.5$ and $w_{\text{geo}}=w_{\text{app}}=w_{\text{sem}}=0.5/3$. This configuration prioritizes 2D projection consistency ($w_{\text{iou}}$) to ensure stable tracking under viewpoint changes, while leveraging geometric anchors, appearance discrimination, and semantic compatibility cues in a balanced manner to address complex scenarios such as dense occlusion, class confusion, and label noise.
Specifically, $\mathrm{IoU}$ measures 2D projection consistency by comparing the detection box $\mathbf{b}_{n_k}$ with the map node's projected box $\Pi_t(\mathcal{P}_{v_j})$. $\boldsymbol{\mu}_{v_j}$ represents the centroid of the accumulated point cloud for map node $v_j$, serving as a stable anchor for geometric association. $\mathcal{K}$ denotes a Gaussian kernel $\mathcal{K}(d)=\exp(-d^2/2\sigma^2)$ applied to the Euclidean distance $d=\|\mathbf{x}_{n_k}^{3D}-\boldsymbol{\mu}_{v_j}\|$, mapping the distance to a similarity score within $(0,1]$. The appearance similarity $\cos(\mathbf{h}_{n_k},\mathbf{H}_{v_j})$ compares the color histogram of the current frame with the node's accumulated histogram, enabling effective discrimination even when objects are semantically identical and spatially proximate (\textit{e.g.}, tightly arranged bottles of the same category)~\cite{ok2019robust, hu2025dyo}. Semantic compatibility $\mathcal{J}$ is defined as the cosine similarity between CLIP features, facilitating the association of synonymous expressions in an open-vocabulary setting (\textit{e.g.}, ``handle'' vs. ``drawer handle'').    

To improve matching efficiency and prune false candidates, we only build graph edges between candidate pairs $(n_k, v_j)$ that satisfy the gating conditions: $\|\mathbf{x}_{n_k}^{3D} - \boldsymbol{\mu}_{v_j}\| < \tau_{\text{d}} \land \text{IoU}(\mathbf{b}_{n_k}, \Pi_t(\mathcal{P}_{v_j})) > 0 \land S^{(t)}(n_k \to v_j) \ge \tau_{\text{ass}}$.
Here, $\tau_{\text{d}}= \min(0.15 \, \text{m}, 0.5 \cdot d_{v_j})$, with $d_{v_j}$ being the diagonal length of the bounding box of $v_j$, to achieve a scale-adaptive geometric proximity constraint; $\tau_{\text{ass}}=0.45$ filters low-quality matches.
On this sparse bipartite graph, we solve the maximum-weight one-to-one matching problem:
\begin{equation}
\max_{\mathbf{A}} \ \sum_{k,j} A_{kj} S^{(t)}(n_k \to v_j)
\quad \text{s.t. } A_{kj}\in\{0,1\},\ \sum_j A_{kj}\le 1,\ \sum_k A_{kj}\le 1.
\end{equation}
This matching is non-exhaustive: due to the gating, some nodes may remain unmatched. Successfully matched pairs $(n_k, v_j)$ are used to update the state of the map node instance $v_j$ (\textit{e.g.}, fusing the new observation point cloud into $\mathcal{P}$, recomputing the centroid $\boldsymbol{\mu}$, and updating the appearance $\mathbf{H}$ via exponential moving average); unmatched detections $n_k$ are initialized as new map nodes.

\noindent\textbf{Edge Association.}
The frame-by-frame evidence provided by the 2D layer is sparse, noisy, and lacks temporal consistency. Specifically, under the expanded benchmark, manipulable objects in tabletop scenes exhibit minute scales, and nodes of the same category coexist in close proximity alongside occlusions. In hierarchical scenes, fine-grained nodes (\textit{e.g.}, adjacent drawer panels) not only suffer from severe appearance homogenization, but their physical boundaries are also difficult to clearly define across various viewpoints. These issues pose significant challenges for determining functional edges within the map: 1) Single-frame visual and geometric cues are highly sensitive to occlusion and viewpoint, rendering hard assignments based on instantaneous scores susceptible to sporadic noise; 2) Fine-grained nodes exhibit multi-candidate ambiguity when facing similar objects (\textit{e.g.}, side-by-side drawers, bottles), and lacking uncertainty modeling, early noise can mislead and lock in erroneous associations; 3) Perturbations in instantaneous evidence can cause frequent association switching, destabilizing the functional topology. To address these issues, we model edge association as a time-evolving graph optimization problem, robustly determining functional edges under multi-candidate ambiguity and noisy observations through joint design of evidence accumulation, entropy regularization, and temporal smoothing.

Specifically, for an $f \in \mathcal{C}\cup\mathcal{U}$, we first construct its object candidate set $\mathcal{O}_f$ based on 2D functional edge candidates and valid 3D reconstruction conditions. 
Let $z_{o \leftarrow f}^{(t)}$ denote the soft assignment probability of associating $f$ with a candidate object $o \in \mathcal{O}_f$ at time $t$, and let $\mathbf{z}_f^{(t)}$ be the distribution over $\mathcal{O}_f$, satisfying $\sum_{o} z_{o \leftarrow f}^{(t)} = 1$ and $z_{o \leftarrow f}^{(t)} \ge 0$. 
We optimize the soft assignment distributions for all fine-grained nodes at each time $t$ by maximizing the objective function:
\begin{equation}
\max_{\mathbf{z}_f^{(t)}} \ \bigg[ \sum_{o \in \mathcal{O}_f} z_{o \leftarrow f}^{(t)} L^{1:t}(o \leftarrow f) + \mathcal{H}(\mathbf{z}_f^{(t)}) - \mathcal{D}(\mathbf{z}_f^{(t)}, \mathbf{z}_f^{(t-1)}) \bigg].
\end{equation}
This objective function balances three aspects. The first term is the data fidelity term, at the core of which is the historically accumulated evidence $L^{1:t}$. To address the issues of sparse and noisy evidence, we employ log-odds accumulation to fuse observations across time:
\begin{equation}
L^{1:t}(o \leftarrow f) := \sum_{\tau \in \mathcal{T}_f^{t}(o)} \mathrm{logit}\big(s_{2D}^{(\tau)}(o \leftarrow f)\big), 
\end{equation}
where $\mathrm{logit}(x)=\log\frac{x}{1-x}$, and $\mathcal{T}_f^{t}(o)$ is the set of frame indices up to $t$ where the edge $(o \leftarrow f)$ was validly observed (\textit{i.e.}, $o \in \mathcal{O}_f$). This method leverages long-term observations to offset instantaneous noise, widening the evidence gap between strong and weak candidates over time.

The second term is the entropy regularization term $\mathcal{H}(\mathbf{z}_f^{(t)})$, used to suppress premature polarization:
\begin{equation}
\mathcal{H}(\mathbf{z}_f^{(t)}) = - \sum_{o \in \mathcal{O}_f} z_{o \leftarrow f}^{(t)} \log z_{o \leftarrow f}^{(t)}.
\end{equation}
This term encourages the distribution to maintain uncertainty when evidence is still insufficient, avoiding locking onto a single association when evidence is ambiguous.

The third term is the temporal smoothing term $\mathcal{D}(\cdot, \cdot)$, used to suppress association jitter:
\begin{equation}
\mathcal{D}(\mathbf{z}_f^{(t)}, \mathbf{z}_f^{(t-1)}) = \frac{1}{2}\sum_{o \in \mathcal{O}_f} (z_{o \leftarrow f}^{(t)} - z_{o \leftarrow f}^{(t-1)})^2.
\end{equation}
This term penalizes drastic changes in the soft assignment distribution between adjacent frames, ensuring the association updates only when new evidence is sufficiently strong.

This objective defines a graph optimization problem over the candidate object set $\mathcal{O}_f$ for the fine-grained node $f$. 
As the final determination of a functional edge is essentially a discrete choice from a finite set of candidate objects, we construct an integrated decision score:
\begin{equation}
\Lambda^{(t)}(o \leftarrow f) := L^{1:t}(o \leftarrow f) + \log z_{o \leftarrow f}^{(t)},
\end{equation}
and for each $f$, we take the following over the $\mathcal{O}_f$:
\begin{equation}
\tilde{o}_f^{(t)} := \arg\max_{o \in \mathcal{O}_f} \Lambda^{(t)}(o \leftarrow f),
\end{equation}
\textit{\textit{i.e.},} we adopt a Top-1 strategy to select the object with the highest score, incorporating $(\tilde{o}_f^{(t)} \leftarrow f)$ into $\mathcal{E}$ as the functional edge at time $t$. 

\subsection{Global Hierarchy Shaping}\label{Meth:GHS}

As described in Section~\ref{Meth:2DFEG}, directly parsing the multi-level structure at the 2D layer leads to candidate combinatorial complexity and single-frame view ambiguity.
Whereas in the 3D layer, nodes and functional edges have completed cross-frame fusion via graph optimization, 2D noise is significantly suppressed, and reliable evidence can be provided using precise 3D geometry.
Therefore, multi-hierarchical relationships are better determined in 3D.

Given the result from the last step $\mathcal{F}_o = \{ f \in \mathcal{F} \mid (o \leftarrow f) \in \mathcal{E} \}$ for each object $o \in \mathcal{O}$, the goal of global hierarchy shaping is to rewrite suitable $o \leftarrow u$ edges among them into hierarchical $o \leftarrow c \leftarrow u$ connections.
We leverage the hierarchical role feasibility $r_f^{\mathcal{C}}(o), r_f^{\mathcal{U}}(o) \in \{0, 1\}$ provided by $\mathcal{M}_t$ to define the candidate functional carrier set $\mathcal{C}_o = \{f \in \mathcal{F}_o \mid r_f^{\mathcal{C}}(o) = 1\}$ and the candidate interactive unit set $\mathcal{U}_o = \{f \in \mathcal{F}_o \mid r_f^{\mathcal{U}}(o) = 1\}$ within $\mathcal{F}_o$. 
A one-time pairing is performed on $\mathcal{C}_o \times \mathcal{U}_o$ by defining binary pairing variables $Y_{c \leftarrow u} \in \{0, 1\}$ where $\sum_{c \in \mathcal{C}_o} Y_{c \leftarrow u} \le 1 \quad (\forall u \in \mathcal{U}_o)$, ensuring that each interactive unit $u$ is assigned to at most one functional carrier $c$.
The pairing objective is to maximize a score based on semantics and geometry:
\begin{equation}
\begin{split}
\max_{\{Y_{c \leftarrow u}\}} \sum_{u \in \mathcal{U}_o} \sum_{c \in \mathcal{C}_o} & Y_{c \leftarrow u} \left( C_{\text{prior}}(c, u) + G_{\text{near}}(c, u) \right).
\end{split}
\end{equation}
Here, $C_{\text{prior}}(c, u)$ is the semantic compatibility defined by $\mathcal{M}_t$; $G_{\text{near}}(c, u)$ is the 3D geometric proximity score.
Finally, we restructure the graph based on the optimization result $Y_{c \leftarrow u}$: if $Y_{c \leftarrow u}=1$, the $o \leftarrow u$ edge is removed and $c \leftarrow u$ is added to construct the $o \leftarrow c \leftarrow u$ hierarchy; if $\sum_c Y_{c \leftarrow u}=0$, the direct $o \leftarrow u$ connection is retained.

\section{Datasets and Benchmarks}
\begin{figure}[t]
  \centering
   \includegraphics[width=1.0\linewidth]{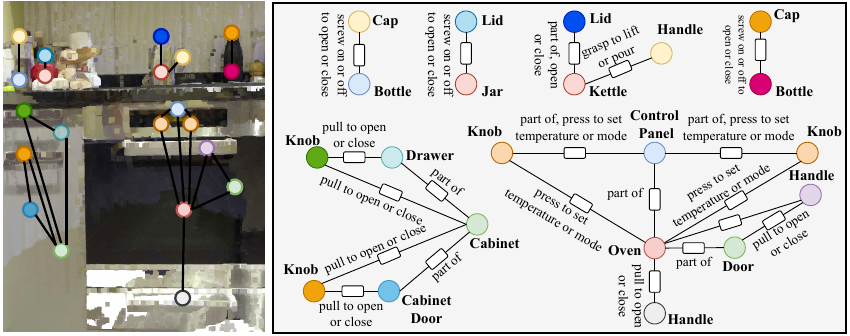}
   \caption{Examples of the improved benchmark. We introduce tabletop manipulable objects and hierarchical relationships.}
   \label{fig:data}
\end{figure}

OpenFunGraph \cite{zhang2025open} provides a coarse-grained benchmark on functional 3D scene graphs on FunGraph3D and modified SceneFun3D, with most annotations focusing on large furniture (\textit{e.g.} cabinet, refrigerators), linking all interactive elements to the object without considering natural hierarchies (\textit{e.g.} wardrobe-drawer-knob).
As shown in Figure~\ref{fig:data}, our improved benchmark makes it more fine-grained and closer to realistic setting by introducing (1) various tabletop objects which play a common role in manipulation tasks (\textit{e.g.} kettle, bottle); (2) hierarchical relations which indicate the specific functional carrier of each interactive element (\textit{e.g.} pot-lid-knob, chest-drawer-knob). 
To enable it, we mimic the annotation tool in OpenFunGraph and extend it with hierarchical annotation features.
Annotators can navigate the 3D scene and annotate the instances of object - functional carrier - interactive unit triplets with a free-form label. 
To enable the above extensions, we conducted a systematic hierarchical functional annotation and statistical analysis on FunGraph3D \cite{zhang2025open} and SceneFun3D \cite{delitzas2024scenefun3d}. 
FunGraph3D contains 722 nodes ($\mathcal{O}=224$, $\mathcal{C}=94$, $\mathcal{U}=404$), with 592 functional edges in total; among them, 118 are hierarchical relations, and 140 belong to the tabletop category. 
In contrast, we find that the scenes in SceneFun3D do not include many tabletop items.
Therefore, it is just used for evaluating hierarchical performance.
It contains 417 nodes ($\mathcal{O}=106$, $\mathcal{C}=90$, $\mathcal{U}=221$), with 438 functional edges; among them, 120 are hierarchical relations.

\section{Experiments}

\subsection{Experimental Setup}
\noindent\textbf{Datasets and Metrics.}
We evaluate functional 3D scene graphs on the extended-annotation SceneFun3D and FunGraph3D datasets. 
The evaluation protocol follows the retrieval-based setting of OpenFunGraph~\cite{zhang2025open}, but with two key modifications to accommodate the semantic granularity differences and dense local label distribution introduced by free-text labels and our tabletop/multi-level annotation extensions.
First, in node/triplet evaluation, a \texttt{hit} is defined as Recall@3/Recall@5, or a similarity score exceeding 0.75/0.70, to mitigate false negatives caused by subtle phrasing differences in synonymous labels.
Second, the Hungarian algorithm is used to perform optimal one-to-one matching between ground-truth nodes and predicted nodes, preventing a single predicted node or edge from being \texttt{hit} by multiple close ground-truths, which would inflate the recall rate. Apart from these two points, the rest of the procedure remains consistent with OpenFunGraph.
Considering our introduction of tabletop objects and hierarchical structures, node metrics are reported separately for three levels: objects, functional carriers, and interactive units, with the performance on the tabletop subset tallied separately. 
For triplet evaluation, in addition to overall results, we also report performance on the hierarchical and tabletop subsets.

\noindent\textbf{Baselines.}
We primarily compare our method against two representative approaches: FunGraph \cite{rotondi2025fungraph} and OpenFunGraph \cite{zhang2025open}. 
FunGraph proposed a functional-aware 3D scene graph, establishing \texttt{has-part} connections for objects and functional parts in a three-stage pipeline using 2D-trained functional part detection and VLM contextual annotation.
OpenFunGraph provides a baseline closer to our task setting for inferring functional 3D scene graphs in an open-vocabulary manner.
To avoid performance discrepancies caused by the language model, the GPT-4~\cite{achiam2023gpt} used in the original OpenFunGraph was replaced with GPT-5~\cite{openai_gpt5_system_card_2025}, the model used in this study.
However, OpenFunGraph faces difficulties in complex scenarios that involve tabletop objects and hierarchical relationships. 
Therefore, while maintaining its simple recognize-then-connect paradigm, we incorporate additional prompts into both its detection module and the LLM reasoning process.
By retaining tabletop categories in its detection's semantic filter and providing the LLM with systematic examples of tabletop functions, we implemented \textit{OpenFunGraph Tabletop}. 
Building on this, we further incorporate relevant examples and instructions for constructing hierarchical relationships into the LLM prompts, resulting in \textit{OpenFunGraph Hierarchy}. 

\noindent\textbf{Data processing.}
Achieving fine-grained scene understanding inevitably leads to a surge in information density. For instance, on the FunGraph3D dataset, the number of nodes detected and recalled by our system is 2.1 and 1.8 times that of OpenFunGraph, respectively. 
To control computational overhead, we apply keyframe downsampling to all input sequences. This strategy compresses our runtime to be comparable with the baseline.
All subsequently reported metrics for our method are based on these downsampled sequences. The ability to maintain an extremely high recall rate under such sparse observations is attributed to the reliable anchoring of 2D functional edges based on concrete visual evidence, alongside the robust multi-cue association and graph optimization mechanisms, whose historical accumulation and temporal smoothing effectively bridge the temporal gaps caused by downsampling.

\subsection{Results}
\noindent\textbf{Quantitative Comparison.}
The quantitative results are shown in Table~\ref{tab:node_eval} and Table~\ref{tab:trip_eval}.
Since SceneFun3D does not contain tabletop objects at the node level, we get a similar node retrieval ability with OpenFunGraph on easily detected large furniture.
However, we consistently outperform on functional carriers for our designated process on that hierarchy. 
On the FunGraph3D dataset, which features numerous tabletop objects and hierarchical structures, we demonstrate significant advantages across all node evaluation metrics.
To be specific in each baseline, FunGraph relies on 2D-trained part detection and contextual annotation, lacking open-vocabulary capabilities, cross-frame fusion, and explicit hierarchical modeling. 
It struggles to form stable grounding for dense small parts and multi-level relationships, thus exhibiting overall lower node and triplet metrics on both datasets.
OpenFunGraph relies solely on geometric proximity and CLIP mean aggregation, making it difficult to distinguish small-scale, texture-less tabletop nodes and densely arranged fine-grained hierarchical nodes (\textit{e.g.}, bottles, drawers) during 3D fusion.
Overall, the improvement in our node recall is attributed to: 1) the dynamic Interactability Map, which expands the functional node candidate space; 2) more robust cross-frame node association; and 3) building upon 2D functional anchoring, the carrier role is explicitly preserved via temporal association and 3D global hierarchy shaping.

At the triplet level, the significant discrepancy between the Tabletop and Hierarchy subsets corroborates the challenges regarding the lack of visual anchoring and attribution uncertainty.
FunGraph~\cite{rotondi2025fungraph} relies on detection box overlap and incremental fusion; however, due to the lack of global temporal optimization, it struggles to rectify single-view occlusions, false detections, and accumulated errors.
OpenFunGraph falters under noise and aliasing due to the lack of explicit modeling for tabletop and hierarchical structures and its reliance on coarse 3D spatial overlap to filter candidates. 
OpenFunGraph Hierarchy still adheres to the geometric proximity paradigm, unable to resolve bounding box aliasing among spatially adjacent nodes of the same category (\textit{e.g.}, dense drawers); meanwhile, OpenFunGraph Tabletop is plagued by the geometric drift of tiny nodes and attribution ambiguity of interactive units caused by a lack of temporal consistency. In contrast, we leverage 2D visual anchoring and temporal optimization to robustly determine functional relationships.

\begin{table*}[t]
\setlength{\tabcolsep}{4px}
\centering
\caption{Node evaluation on the SceneFun3D and FunGraph3D datasets. ``--'' indicates SceneFun3D does not contain Tabletop Nodes.}
\resizebox{\linewidth}{!}{
\begin{tabular}{l cccc c ccccc}
\toprule
& \multicolumn{4}{c}{\textbf{SceneFun3D}~\cite{delitzas2024scenefun3d}}
&& \multicolumn{5}{c}{\textbf{FunGraph3D}~\cite{zhang2025open}}
\\
\cmidrule{2-5}  \cmidrule{7-11}
\textbf{Methods} & Objects
& Fun. Carriers
& Inter. Elements
& Overall Nodes
&
& Objects
& Fun. Carriers
& Inter. Elements
& Tabletop Nodes
& Overall Nodes
\\
\midrule
FunGraph
& 47.2 & 32.2 & 55.2 & 48.2
&
& 28.6 & 26.6 & 20.3 & 25.7 & 23.7 \\
OpenFunGraph
& 72.6 & 13.3 & 68.8 & 57.8
&
& 49.1 & 11.7 & 28.7 & 39.3 & 32.8 \\
OpenFunGraph Tabletop
& - & - & - & -
&
& 53.5 & 11.7 & 42.1 & 52.1 & 41.7 \\
OpenFunGraph Hierarchy
& 72.6 & 47.8 & 68.8 & 65.2
&
& 53.5 & 41.5 & 43.3 & 55.7 & 46.3 \\
Ours
& 75.5 & 66.7 & 70.1 & 70.7
&
& 56.3 & 78.7 & 58.2 & 71.4 & 60.2 \\
\bottomrule
\end{tabular}
}
\label{tab:node_eval}
\end{table*}


\begin{table}[t]
\centering
\begin{minipage}[t]{0.49\textwidth}
\centering
\setlength{\tabcolsep}{4pt} 
\caption{Triplet evaluation on the SceneFun3D and FunGraph3D. ``$\times$'' indicates the method fails to recall the triplets.}
\vspace{-2mm}
\resizebox{\linewidth}{!}{
\begin{tabular}{l cc c ccc}
\toprule
& \multicolumn{2}{c}{\textbf{SceneFun3D}~\cite{delitzas2024scenefun3d}}
&& \multicolumn{3}{c}{\textbf{FunGraph3D}~\cite{zhang2025open}}
\\
\cmidrule{2-3}  \cmidrule{5-7}
\textbf{Methods} 
& Hierarchical 
& Overall 
&
& Hierarchical 
& Tabletop 
& Overall 
\\
\midrule
FunGraph~\cite{rotondi2025fungraph}
& $\times$ & 23.5
&
& $\times$ & 6.4 & 19.1 \\
OpenFunGraph~\cite{zhang2025open} 
& $\times$ & 30.8
&
& $\times$ & 13.6 & 21.8 \\
OpenFunGraph Tabletop 
& - & -
&
& $\times$ & 42.1 & 28.5 \\
OpenFunGraph Hierarchy
& 12.5 & 37.6
&
& 16.1 & 42.9 & 35.1 \\
Ours
& 52.5 & 50.9
&
& 61.9 & 61.4 & 55.6 \\
\bottomrule
\end{tabular}
}
\label{tab:trip_eval}
\end{minipage}\hfill
\begin{minipage}[t]{0.49\textwidth}
\centering
\setlength{\tabcolsep}{4pt}
\small
\caption{Ablations on FunGraph3D \cite{zhang2025open} dataset for nodes and triplets.}
\vspace{-2mm}
\resizebox{\linewidth}{!}{
\begin{tabular}{l cc ccc}
\toprule
& \multicolumn{2}{c}{\textbf{Nodes}} & \multicolumn{3}{c}{\textbf{Triplets}} \\
\cmidrule(lr){2-3}\cmidrule(lr){4-6}
\textbf{Methods} & Table-top & Overall & Hier. & Table-top & Overall \\
\midrule
Assoc-OpenFG     & 58.1 & 46.1 & 46.6 & 48.6 & 40.9 \\
No-GO-Count      & 67.1 & 55.1 & 50.8 & 57.9 & 47.1 \\
2D-Hierarchy     & 70.3 & 51.9 & 44.9 & 60.7 & 49.2 \\
Ours             & 71.4 & 60.2 & 61.9 & 61.4 & 55.6 \\
\bottomrule
\end{tabular}}
\label{tab:nodes_triplets_eval}
\end{minipage}
\end{table}

\begin{figure*}[t]
  \centering
   \includegraphics[width=1.0\linewidth]{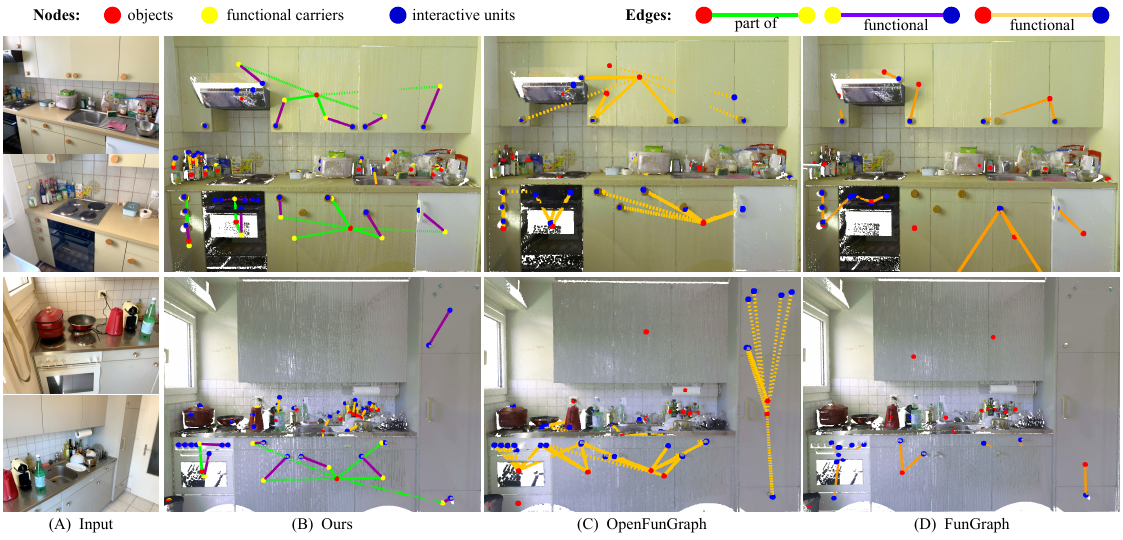}
   \caption{Qualitative results. Our constructed functional 3D scene graph features a hierarchical structure and covers a more comprehensive range of manipulable objects.}
   \label{fig:qualitative}
\end{figure*}

\noindent\textbf{Qualitative Comparison.}
Figure~\ref{fig:qualitative} provides a more intuitive comparison of different methods in real-world scenarios.
OpenFunGraph and FunGraph often only construct a few tabletop-level objects and can hardly recognize their interactive units, let alone reliable functional edges. 
Regarding multi-level relationships, closely arranged similar carriers (\textit{e.g.}, rows of drawers, adjacent cabinet doors) introduce strong ambiguity. 
OpenFunGraph and FunGraph often ignore the functional carriers altogether, instead hard-linking the interactive units directly to the object body; even so, under occlusion and viewpoint changes, this direct connection lacks geometric and topological grounding, and incorrect associations occur frequently.
In contrast, our method can accurately restore typical hierarchical structures, and successfully construct most nodes and functional edges in tabletop scenarios. 
See the supplement for more qualitative results.

\noindent\textbf{Real-World Manipulation Tasks.}
Furthermore, to validate the utility of the constructed hierarchical functional scene graph in real-world manipulation tasks, we conduct multi-task natural language-guided manipulation demonstrations on a physical robotic arm platform; experimental details and results are provided in the supplementary material.

\subsection{Ablation Studies}
\noindent\textbf{Ablation Settings.}
We define three ablations on the FunGraph3D dataset~\cite{zhang2025open} to
evaluate the contribution of each key component:
1) \textit{Assoc-OpenFG} replaces our multi-cued node association with OpenFunGraph's 3D-IoU + CLIP cosine similarity;
2) \textit{No-GO-Count} removes graph optimization, selecting the edge via Top-1 2D observation count.
3) \textit{2D-Hierarchy} directly parses the hierarchical relationship at the 2D level without refinement at the 3D level;

\noindent\textbf{Results and Analysis.}
As shown in Table~\ref{tab:nodes_triplets_eval},
\textit{Assoc-OpenFG} is significantly impaired: 3D-IoU has poor discriminability for adjacent similar instances (\textit{e.g.}, bottles, drawers) and CLIP is unstable for parts, while our multi-cued scoring effectively suppresses such mismatches.
\textit{No-GO-Count} shows a decrease: simple counting cannot resolve 2D ambiguities, indicating frequency alone is insufficient to replace energy-based temporal graph optimization.
\textit{2D-Hierarchy} declines primarily in hierarchical triplets: parsing the $\mathcal{O} \leftarrow \mathcal{C} \leftarrow \mathcal{U}$ hierarchy in 2D introduces candidate explosion and view ambiguity, while moving this determination to 3D avoids these issues.
See the supplement for more ablations.

\section{Conclusion}



This paper aims to address the deficiencies of existing functional 3D scene graphs in handling tabletop objects and hierarchical structures for fine-grained manipulation. We first extend the benchmark to cover tabletop manipulable objects and hierarchical structures. 
To tackle the challenges introduced by this expansion, which existing paradigms struggle to address, including the lack of visual anchoring in relational reasoning, instance confusion during cross-frame fusion, and attribution uncertainty under dynamic viewpoints, we propose an open-vocabulary, visually grounded and graph-optimization-based pipeline.
This approach anchors functional edges using 2D evidence, stabilizes instance identity via multi-cue association in 3D, robustly determines edge attribution through temporal graph optimization, and finally performs global hierarchy shaping to recover the hierarchical structure.
Experiments demonstrate that our method significantly outperforms existing approaches on the expanded benchmark.

\noindent\textbf{Limitations.} 
1) Ambiguities may persist under extreme occlusion or restricted viewpoints; 2) Performance is bounded by the quality of detection/segmentation and VLM discrimination.
Future work will explore active viewpoint selection.
\section*{Acknowledgements}

\begingroup
\setlength{\emergencystretch}{4em}
\begin{sloppypar}
This work was supported in part by the following funding sources: the National Science and Technology Major Project (Grant 2025ZD\allowbreak{}1606303); the Fundamental and Interdisciplinary Disciplines Breakthrough Plan of the Ministry of Education of China (Grant JYB\allowbreak{}2025\allowbreak{}XDXM\allowbreak{}503); and the National Science Foundation of China (Grant 62495092). 
Additional support for this work was provided by the Swiss National Science Foundation Advanced Grant 216260, ``Beyond Frozen Worlds: Capturing Functional 3D Digital Twins from the Real World''; and by the European Union's Horizon Europe research and innovation programme under grant agreement number 101214398 (ELLIOT), fully funded by the Swiss State Secretariat for Education, Research and Innovation (SERI). 
Alexandros Delitzas is also supported by a doctoral fellowship from the Max Planck ETH Center for Learning Systems (CLS).
\end{sloppypar}
\endgroup

%
%
\bibliographystyle{splncs04}
\bibliography{main}

@String(IJCV  = {Int. J. Comput. Vis.})

@String(CVPR  = {IEEE Conf. Comput. Vis. Pattern Recog.})

@String(ICCV  = {Int. Conf. Comput. Vis.})

@String(ECCV  = {Eur. Conf. Comput. Vis.})

@String(NeurIPS = {Adv. Neural Inform. Process. Syst.})

@String(ICLR  = {Int. Conf. Learn. Represent.})

@String(BMVC  = {Brit. Mach. Vis. Conf.})

@String(CVPRW = {IEEE Conf. Comput. Vis. Pattern Recog. Worksh.})

@String(TOG   = {ACM Trans. Graph.})

@String(ACMMM = {ACM Int. Conf. Multimedia})

@String(IJCV  = {IJCV})

@String(CVPR  = {CVPR})

@String(ICCV  = {ICCV})

@String(ECCV  = {ECCV})

@String(NeurIPS = {NeurIPS})

@String(ICLR  = {ICLR})

@String(BMVC  =	{BMVC})

@String(CVPRW = {CVPRW})

@String(TOG   = {ACM TOG})

@String(ACMMM = {ACM MM})

@String(IJCV = {Int. J. Comput. Vis.})

@String(CVPR= {IEEE Conf. Comput. Vis. Pattern Recog.})

@String(ICCV= {Int. Conf. Comput. Vis.})

@String(ECCV= {Eur. Conf. Comput. Vis.})

@String(BMVC= {Brit. Mach. Vis. Conf.})

@String(TOG= {ACM Trans. Graph.})

@String(ACMMM= {ACM Int. Conf. Multimedia})

@String(ICLR = {Int. Conf. Learn. Represent.})

@String(CVPRW= {IEEE Conf. Comput. Vis. Pattern Recog. Worksh.})

@String(CVPRW= {CVPRW})

@inproceedings{zhang2021exploiting,
  title={Exploiting edge-oriented reasoning for 3d point-based scene graph analysis},
  author={Zhang, Chaoyi and Yu, Jianhui and Song, Yang and Cai, Weidong},
  booktitle=cvpr,
  year={2021}
}

@inproceedings{koch2024open3dsg,
  title={{Open3dsg: Open-vocabulary 3d scene graphs from point clouds with queryable objects and open-set relationships}},
  author={Koch, Sebastian and Vaskevicius, Narunas and Colosi, Mirco and Hermosilla, Pedro and Ropinski, Timo},
  booktitle=cvpr,
  year={2024}
}

@article{li2021ifr,
  title={{IFR-Explore: Learning inter-object functional relationships in 3d indoor scenes}},
  author={Li, Qi and Mo, Kaichun and Yang, Yanchao and Zhao, Hang and Guibas, Leonidas},
  journal=iclr,
  year={2022}
}

@inproceedings{wald2020learning,
  title={{Learning 3d semantic scene graphs from 3d indoor reconstructions}},
  author={Wald, Johanna and Dhamo, Helisa and Navab, Nassir and Tombari, Federico},
  booktitle=cvpr,
  year={2020}
}

@inproceedings{gu2024conceptgraphs,
  title={{ConceptGraphs: Open-vocabulary 3d scene graphs for perception and planning}},
  author={Gu, Qiao and Kuwajerwala, Ali and Morin, Sacha and Jatavallabhula, Krishna Murthy and Sen, Bipasha and Agarwal, Aditya and Rivera, Corban and Paul, William and Ellis, Kirsty and Chellappa, Rama and others},
  booktitle=icra,
  year={2024},
}

@inproceedings{armeni20193d,
  title={{3d scene graph: A structure for unified semantics, 3d space, and camera}},
  author={Armeni, Iro and He, Zhi-Yang and Gwak, JunYoung and Zamir, Amir R and Fischer, Martin and Malik, Jitendra and Savarese, Silvio},
  booktitle=iccv,
  year={2019}
}

@inproceedings{chen2024clip,
  title={CLIP-driven open-vocabulary 3d scene graph generation via cross-modality contrastive learning},
  author={Chen, Lianggangxu and Wang, Xuejiao and Lu, Jiale and Lin, Shaohui and Wang, Changbo and He, Gaoqi},
  booktitle=cvpr,
  year={2024}
}

@inproceedings{zhang2024recognize,
  title={Recognize anything: A strong image tagging model},
  author={Zhang, Youcai and Huang, Xinyu and Ma, Jinyu and Li, Zhaoyang and Luo, Zhaochuan and Xie, Yanchun and Qin, Yuzhuo and Luo, Tong and Li, Yaqian and Liu, Shilong and others},
  booktitle=cvpr,
  year={2024}
}

@misc{liu2024llava,
  title={{Llava-next: Improved reasoning, ocr, and world knowledge}},
  author={Liu, Haotian and Li, Chunyuan and Li, Yuheng and Li, Bo and Zhang, Yuanhan and Shen, Sheng and Lee, Yong Jae},
  year={2024}
}

@inproceedings{liu2024improved,
  title={{Improved baselines with visual instruction tuning}},
  author={Liu, Haotian and Li, Chunyuan and Li, Yuheng and Lee, Yong Jae},
  booktitle=cvpr,
  year={2024}
}

@inproceedings{liu2024visual,
  title={{Visual instruction tuning}},
  author={Liu, Haotian and Li, Chunyuan and Wu, Qingyang and Lee, Yong Jae},
  booktitle=neurips,
  year={2023}
}

@inproceedings{delitzas2024scenefun3d,
  title={Scenefun3d: Fine-grained functionality and affordance understanding in 3d scenes},
  author={Delitzas, Alexandros and Takmaz, Ayca and Tombari, Federico and Sumner, Robert and Pollefeys, Marc and Engelmann, Francis},
  booktitle=cvpr,
  year={2024}
}

@article{atzmon2018point,
  title={Point convolutional neural networks by extension operators},
  author={Atzmon, Matan and Maron, Haggai and Lipman, Yaron},
  journal=tog,
  year={2018}
}

@inproceedings{choy20194d,
  title={4d spatio-temporal convnets: Minkowski convolutional neural networks},
  author={Choy, Christopher and Gwak, JunYoung and Savarese, Silvio},
  booktitle=cvpr,
  year={2019}
}

@inproceedings{hu2021vmnet,
  title={Vmnet: Voxel-mesh network for geodesic-aware 3d semantic segmentation},
  author={Hu, Zeyu and Bai, Xuyang and Shang, Jiaxiang and Zhang, Runze and Dong, Jiayu and Wang, Xin and Sun, Guangyuan and Fu, Hongbo and Tai, Chiew-Lan},
  booktitle=iccv,
  year={2021}
}

@inproceedings{hua2018pointwise,
  title={Pointwise convolutional neural networks},
  author={Hua, Binh-Son and Tran, Minh-Khoi and Yeung, Sai-Kit},
  booktitle=cvpr,
  year={2018}
}

@article{huang2023segment3d,
  title={Segment3d: Learning fine-grained class-agnostic 3d segmentation without manual labels},
  author={Huang, Rui and Peng, Songyou and Takmaz, Ayca and Tombari, Federico and Pollefeys, Marc and Song, Shiji and Huang, Gao and Engelmann, Francis},
  journal=eccv,
  year={2024}
}

@inproceedings{landrieu2018large,
  title={Large-scale point cloud semantic segmentation with superpoint graphs},
  author={Landrieu, Loic and Simonovsky, Martin},
  booktitle=cvpr,
  year={2018}
}

@article{li2018pointcnn,
  title={Pointcnn: Convolution on x-transformed points},
  author={Li, Yangyan and Bu, Rui and Sun, Mingchao and Wu, Wei and Di, Xinhan and Chen, Baoquan},
  journal=neurips,
  year={2018}
}

@inproceedings{qi2017pointnet,
  title={Pointnet: Deep learning on point sets for 3d classification and segmentation},
  author={Qi, Charles R and Su, Hao and Mo, Kaichun and Guibas, Leonidas J},
  booktitle=cvpr,
  year={2017}
}

@article{qi2017pointnet++,
  title={Pointnet++: Deep hierarchical feature learning on point sets in a metric space},
  author={Qi, Charles Ruizhongtai and Yi, Li and Su, Hao and Guibas, Leonidas J},
  journal=neurips,
  year={2017}
}

@inproceedings{thomas2019kpconv,
  title={Kpconv: Flexible and deformable convolution for point clouds},
  author={Thomas, Hugues and Qi, Charles R and Deschaud, Jean-Emmanuel and Marcotegui, Beatriz and Goulette, Fran{\c{c}}ois and Guibas, Leonidas J},
  booktitle=iccv,
  year={2019}
}

@inproceedings{weder2024labelmaker,
  title={LabelMaker: Automatic semantic label generation from RGB-D trajectories},
  author={Weder, Silvan and Blum, Hermann and Engelmann, Francis and Pollefeys, Marc},
  booktitle=threedv,
  year={2024}
}

@inproceedings{engelmann20203d,
  title={3d-mpa: Multi-proposal aggregation for 3d semantic instance segmentation},
  author={Engelmann, Francis and Bokeloh, Martin and Fathi, Alireza and Leibe, Bastian and Nie{\ss}ner, Matthias},
  booktitle=cvpr,
  year={2020}
}

@inproceedings{han2020occuseg,
  title={Occuseg: Occupancy-aware 3d instance segmentation},
  author={Han, Lei and Zheng, Tian and Xu, Lan and Fang, Lu},
  booktitle=cvpr,
  year={2020}
}

@inproceedings{hou20193d,
  title={3d-sis: 3d semantic instance segmentation of rgb-d scans},
  author={Hou, Ji and Dai, Angela and Nie{\ss}ner, Matthias},
  booktitle=cvpr,
  year={2019}
}

@inproceedings{jiang2020pointgroup,
  title={Pointgroup: Dual-set point grouping for 3d instance segmentation},
  author={Jiang, Li and Zhao, Hengshuang and Shi, Shaoshuai and Liu, Shu and Fu, Chi-Wing and Jia, Jiaya},
  booktitle=cvpr,
  year={2020}
}

@inproceedings{schult2023mask3d,
  title={Mask3d: Mask transformer for 3d semantic instance segmentation},
  author={Schult, Jonas and Engelmann, Francis and Hermans, Alexander and Litany, Or and Tang, Siyu and Leibe, Bastian},
  booktitle=icra,
  year={2023}
}

@inproceedings{vu2022softgroup,
  title={Softgroup for 3d instance segmentation on point clouds},
  author={Vu, Thang and Kim, Kookhoi and Luu, Tung M and Nguyen, Thanh and Yoo, Chang D},
  booktitle=cvpr,
  year={2022}
}

@article{yue2023agile3d,
  title={AGILE3d: Attention guided interactive multi-object 3d segmentation},
  author={Yue, Yuanwen and Mahadevan, Sabarinath and Schult, Jonas and Engelmann, Francis and Leibe, Bastian and Schindler, Konrad and Kontogianni, Theodora},
  journal=iclr,
  year={2024}
}

@article{engelmann2024opennerf,
  title={OpenNeRF: Open set 3d neural scene segmentation with pixel-wise features and rendered novel views},
  author={Engelmann, Francis and Manhardt, Fabian and Niemeyer, Michael and Tateno, Keisuke and Pollefeys, Marc and Tombari, Federico},
  journal=iclr,
  year={2024}
}

@inproceedings{kerr2023lerf,
  title={Lerf: Language embedded radiance fields},
  author={Kerr, Justin and Kim, Chung Min and Goldberg, Ken and Kanazawa, Angjoo and Tancik, Matthew},
  booktitle=iccv,
  year={2023}
}

@article{jatavallabhula2023conceptfusion,
  title={Conceptfusion: Open-set multimodal 3d mapping},
  author={Jatavallabhula, Krishna Murthy and Kuwajerwala, Alihusein and Gu, Qiao and Omama, Mohd and Chen, Tao and Maalouf, Alaa and Li, Shuang and Iyer, Ganesh and Saryazdi, Soroush and Keetha, Nikhil and others},
  journal={ICRA2023 Workshop on Pretraining for Robotics (PT4R)},
  year={2023}
}

@inproceedings{peng2023openscene,
  title={Openscene: 3d scene understanding with open vocabularies},
  author={Peng, Songyou and Genova, Kyle and Jiang, Chiyu and Tagliasacchi, Andrea and Pollefeys, Marc and Funkhouser, Thomas and others},
  booktitle=cvpr,
  year={2023}
}

@article{takmaz2023openmask3d,
  title={Openmask3d: Open-vocabulary 3d instance segmentation},
  author={Takmaz, Ay{\c{c}}a and Fedele, Elisabetta and Sumner, Robert W and Pollefeys, Marc and Tombari, Federico and Engelmann, Francis},
  journal=neurips,
  year={2023}
}

@inproceedings{zhou2024feature,
  title={Feature 3dGS: Supercharging 3d gaussian splatting to enable distilled feature fields},
  author={Zhou, Shijie and Chang, Haoran and Jiang, Sicheng and Fan, Zhiwen and Zhu, Zehao and Xu, Dejia and Chari, Pradyumna and You, Suya and Wang, Zhangyang and Kadambi, Achuta},
  booktitle=cvpr,
  year={2024}
}

@article{zuo2024fmgs,
  title={Fmgs: Foundation model embedded 3d gaussian splatting for holistic 3d scene understanding},
  author={Zuo, Xingxing and Samangouei, Pouya and Zhou, Yunwen and Di, Yan and Li, Mingyang},
  journal=ijcv,
  year={2024}
}

@inproceedings{qin2024langsplat,
  title={Langsplat: 3d language gaussian splatting},
  author={Qin, Minghan and Li, Wanhua and Zhou, Jiawei and Wang, Haoqian and Pfister, Hanspeter},
  booktitle=cvpr,
  year={2024}
}

@inproceedings{huang2022multi,
  title={Multi-view transformer for 3d visual grounding},
  author={Huang, Shijia and Chen, Yilun and Jia, Jiaya and Wang, Liwei},
  booktitle=cvpr,
  year={2022}
}

@inproceedings{yang2021sat,
  title={Sat: 2d semantics assisted training for 3d visual grounding},
  author={Yang, Zhengyuan and Zhang, Songyang and Wang, Liwei and Luo, Jiebo},
  booktitle=iccv,
  year={2021}
}

@inproceedings{hsu2023ns3d,
  title={Ns3d: Neuro-symbolic grounding of 3d objects and relations},
  author={Hsu, Joy and Mao, Jiayuan and Wu, Jiajun},
  booktitle=cvpr,
  year={2023}
}

@inproceedings{zhang2023multi3drefer,
  title={Multi3drefer: Grounding text description to multiple 3d objects},
  author={Zhang, Yiming and Gong, ZeMing and Chang, Angel X},
  booktitle=iccv,
  year={2023}
}

@inproceedings{roh2022languagerefer,
  title={Languagerefer: Spatial-language model for 3d visual grounding},
  author={Roh, Junha and Desingh, Karthik and Farhadi, Ali and Fox, Dieter},
  booktitle=corl,
  year={2022}
}

@inproceedings{do2018affordancenet,
  title={Affordancenet: An end-to-end deep learning approach for object affordance detection},
  author={Do, Thanh-Toan and Nguyen, Anh and Reid, Ian},
  booktitle=icra,
  year={2018}
}

@inproceedings{fang2018demo2vec,
  title={Demo2vec: Reasoning object affordances from online videos},
  author={Fang, Kuan and Wu, Te-Lin and Yang, Daniel and Savarese, Silvio and Lim, Joseph J},
  booktitle=cvpr,
  year={2018}
}

@inproceedings{nagarajan2019grounded,
  title={Grounded human-object interaction hotspots from video},
  author={Nagarajan, Tushar and Feichtenhofer, Christoph and Grauman, Kristen},
  booktitle=iccv,
  year={2019}
}

@article{yoshida2024text,
  title={Text-driven affordance learning from egocentric vision},
  author={Yoshida, Tomoya and Kurita, Shuhei and Nishimura, Taichi and Mori, Shinsuke},
  journal={arXiv preprint arXiv:2404.02523},
  year={2024}
}

@article{cho2024dense,
  title={Dense hand-object (HO) GraspNet with full grasping taxonomy and dynamics},
  author={Cho, Woojin and Lee, Jihyun and Yi, Minjae and Kim, Minje and Woo, Taeyun and Kim, Donghwan and Ha, Taewook and Lee, Hyokeun and Ryu, Je-Hwan and Woo, Woontack and others},
  journal=eccv,
  year={2024}
}

@article{banerjee2024introducing,
  title={Introducing HOT3d: An egocentric dataset for 3d hand and object tracking},
  author={Banerjee, Prithviraj and Shkodrani, Sindi and Moulon, Pierre and Hampali, Shreyas and Zhang, Fan and Fountain, Jade and Miller, Edward and Basol, Selen and Newcombe, Richard and Wang, Robert and others},
  journal={arXiv preprint arXiv:2406.09598},
  year={2024}
}

@inproceedings{ye2024g,
  title={G-HOP: Generative hand-object prior for interaction reconstruction and grasp synthesis},
  author={Ye, Yufei and Gupta, Abhinav and Kitani, Kris and Tulsiani, Shubham},
  booktitle=cvpr,
  year={2024}
}

@inproceedings{fan2024hold,
  title={HOLD: Category-agnostic 3d reconstruction of interacting hands and objects from video},
  author={Fan, Zicong and Parelli, Maria and Kadoglou, Maria Eleni and Chen, Xu and Kocabas, Muhammed and Black, Michael J and Hilliges, Otmar},
  booktitle=cvpr,
  year={2024}
}

@inproceedings{zhang2024moho,
  title={Moho: Learning single-view hand-held object reconstruction with multi-view occlusion-aware supervision},
  author={Zhang, Chenyangguang and Jiao, Guanlong and Di, Yan and Wang, Gu and Huang, Ziqin and Zhang, Ruida and Manhardt, Fabian and Fu, Bowen and Tombari, Federico and Ji, Xiangyang},
  booktitle=cvpr,
  year={2024}
}

@article{zhang2024ddf,
  title={DDF-HO: Hand-held object reconstruction via conditional directed distance field},
  author={Zhang, Chenyangguang and Di, Yan and Zhang, Ruida and Zhai, Guangyao and Manhardt, Fabian and Tombari, Federico and Ji, Xiangyang},
  journal=neurips,
  year={2023}
}

@inproceedings{ye2022s,
  title={What's in your hands? 3d reconstruction of generic objects in hands},
  author={Ye, Yufei and Gupta, Abhinav and Tulsiani, Shubham},
  booktitle=cvpr,
  year={2022}
}

@inproceedings{chen2022alignsdf,
  title={Alignsdf: Pose-aligned signed distance fields for hand-object reconstruction},
  author={Chen, Zerui and Hasson, Yana and Schmid, Cordelia and Laptev, Ivan},
  booktitle=eccv,
  year={2022}
}

@article{rosinol20203d,
  title={3d dynamic scene graphs: Actionable spatial perception with places, objects, and humans},
  author={Rosinol, Antoni and Gupta, Arjun and Abate, Marcus and Shi, Jingnan and Carlone, Luca},
  journal={Robotics, Science and Systems},
  year={2020}
}

@article{rosinol2021kimera,
  title={Kimera: From SLAM to spatial perception with 3d dynamic scene graphs},
  author={Rosinol, Antoni and Violette, Andrew and Abate, Marcus and Hughes, Nathan and Chang, Yun and Shi, Jingnan and Gupta, Arjun and Carlone, Luca},
  journal=ijrr,
  year={2021}
}

@inproceedings{koch2024lang3dsg,
  title={Lang3dSG: Language-based contrastive pre-training for 3d Scene Graph prediction},
  author={Koch, Sebastian and Hermosilla, Pedro and Vaskevicius, Narunas and Colosi, Mirco and Ropinski, Timo},
  booktitle=threedv,
  year={2024}
}

@inproceedings{wang2023vl,
  title={Vl-sat: Visual-linguistic semantics assisted training for 3d semantic scene graph prediction in point cloud},
  author={Wang, Ziqin and Cheng, Bowen and Zhao, Lichen and Xu, Dong and Tang, Yang and Sheng, Lu},
  booktitle=cvpr,
  year={2023}
}

@inproceedings{wu2021scenegraphfusion,
  title={{SceneGraphFusion}: Incremental 3d scene graph prediction from rgb-d sequences},
  author={Wu, Shun-Cheng and Wald, Johanna and Tateno, Keisuke and Navab, Nassir and Tombari, Federico},
  booktitle=cvpr,
  year={2021}
}

@inproceedings{wu2023incremental,
  title={Incremental 3d semantic scene graph prediction from rgb sequences},
  author={Wu, Shun-Cheng and Tateno, Keisuke and Navab, Nassir and Tombari, Federico},
  booktitle=cvpr,
  year={2023}
}

@article{zhang2021knowledge,
  title={Knowledge-inspired 3d scene graph prediction in point cloud},
  author={Zhang, Shoulong and Hao, Aimin and Qin, Hong and others},
  journal=neurips,
  year={2021}
}

@article{zhai2022one,
  title={One-shot object affordance detection in the wild},
  author={Zhai, Wei and Luo, Hongchen and Zhang, Jing and Cao, Yang and Tao, Dacheng},
  journal=ijcv,
  year={2022}
}

@article{achiam2023gpt,
    title     = {Gpt-4 technical report},
    author    = {Achiam, Josh and Adler, Steven and Agarwal, Sandhini and Ahmad, Lama and Akkaya, Ilge and Aleman, Florencia Leoni and Almeida, Diogo and Altenschmidt, Janko and Altman, Sam and Anadkat, Shyamal and others},
    journal   = {arXiv preprint arXiv:2303.08774},
    year      = {2023}
}

@inproceedings{Parelli_2023_CVPR,
    title     = {{CLIP-Guided Vision-Language Pre-Training for Question Answering in 3D Scenes}},
    author    = {Parelli, Maria and Delitzas, Alexandros and Hars, Nikolas and Vlassis, Georgios and Anagnostidis, Sotirios and Bachmann, Gregor and Hofmann, Thomas},
    booktitle = CVPRW,
    year      = {2023}
}

@inproceedings{Delitzas_2023_BMVC,
    author    = {Alexandros Delitzas and Maria Parelli and Nikolas Hars and Georgios Vlassis and Sotirios-Konstantinos Anagnostidis and Gregor Bachmann and Thomas Hofmann},
    title     = {Multi-CLIP: Contrastive Vision-Language Pre-training for Question Answering tasks in 3D Scenes},
    booktitle = BMVC,
    year      = {2023}
}

@article{takmaz2025search3d,
    title     = {{Search3D: Hierarchical Open-Vocabulary 3D Segmentation}},
    author    = {Takmaz, Ayca and Delitzas, Alexandros and Sumner, Robert W. and Engelmann, Francis and Wald, Johanna and Tombari, Federico},
    journal   = RAL,
    year      = {2025}
}

@inproceedings{opencity3d2025,
  title     = {{OpenCity3D: 3D Urban Scene Understanding with Vision-Language Models}},
  author    = {Bieri, Valentin and Zamboni, Marco and Blumer, Nicolas S. and Chen, Qingxuan and Engelmann, Francis},
  booktitle = WACV,
  year      = {2025},
}

@article{weder2023alster,
  title     = {{Alster: A Local Spatio-temporal Expert for Online 3D Semantic Reconstruction}},
  author    = {Weder, Silvan and Engelmann, Francis and Sch{\"o}nberger, Johannes L and Seki, Akihito and Pollefeys, Marc and Oswald, Martin R},
  journal   = WACV,
  year      = {2023}
}

@article{yilmaz2024opendas,
  title     = {{OpenDAS: Open-Vocabulary Domain Adaptation for 2D and 3D Segmentation}},
  author    = {Yilmaz, Gonca and Peng, Songyou and Pollefeys, Marc and Engelmann, Francis and Blum, Hermann},
  journal   = {arXiv preprint arXiv:2405.20141},
  year      = {2024}
}

@inproceedings{takmaz20233d,
  title     = {{3D Segmentation of Humans in Point Clouds with Synthetic Data}},
  author    = {Takmaz, Ay{\c{c}}a and Schult, Jonas and Kaftan, Irem and Ak{\c{c}}ay, Mertcan and Leibe, Bastian and Sumner, Robert and Engelmann, Francis and Tang, Siyu},
  booktitle = ICCV,
  year      = {2023}
}

@inproceedings{zhang2025open,
  title={Open-vocabulary functional 3d scene graphs for real-world indoor spaces},
  author={Zhang, Chenyangguang and Delitzas, Alexandros and Wang, Fangjinhua and Zhang, Ruida and Ji, Xiangyang and Pollefeys, Marc and Engelmann, Francis},
  booktitle=CVPR,
  year={2025}
}

@article{rotondi2025fungraph,
  title={FunGraph: Functionality Aware 3D Scene Graphs for Language-Prompted Scene Interaction},
  author={Rotondi, Dennis and Scaparro, Fabio and Blum, Hermann and Arras, Kai O},
  booktitle=IROS,
  year={2025}
}

@inproceedings{huang2025open,
  title={Open-set image tagging with multi-grained text supervision},
  author={Huang, Xinyu and Huang, Yi-Jie and Zhang, Youcai and Tian, Weiwei and Feng, Rui and Zhang, Yuejie and Xie, Yanchun and Li, Yaqian and Zhang, Lei},
  booktitle=ACMMM,
  year={2025}
}

@inproceedings{fu2025llmdet,
  title={Llmdet: Learning strong open-vocabulary object detectors under the supervision of large language models},
  author={Fu, Shenghao and Yang, Qize and Mo, Qijie and Yan, Junkai and Wei, Xihan and Meng, Jingke and Xie, Xiaohua and Zheng, Wei-Shi},
  booktitle=CVPR,
  year={2025}
}

@article{ravi2024sam2,
  title={SAM 2: Segment Anything in Images and Videos},
  author={Ravi, Nikhila and Gabeur, Valentin and Hu, Yuan-Ting and Hu, Ronghang and Ryali, Chaitanya and Ma, Tengyu and Khedr, Haitham and R{\"a}dle, Roman and Rolland, Chloe and Gustafson, Laura and Mintun, Eric and Pan, Junting and Alwala, Kalyan Vasudev and Carion, Nicolas and Wu, Chao-Yuan and Girshick, Ross and Doll{\'a}r, Piotr and Feichtenhofer, Christoph},
  year={2024}
}

@inproceedings{ok2019robust,
  title={Robust object-based slam for high-speed autonomous navigation},
  author={Ok, Kyel and Liu, Katherine and Frey, Kris and How, Jonathan P and Roy, Nicholas},
  booktitle={2019 International Conference on Robotics and Automation (ICRA)},
  pages={669--675},
  year={2019},
  organization={IEEE}
}

@article{hu2025dyo,
  title={DYO-SLAM: Visual Localization and Object Mapping in Dynamic Scenes},
  author={Hu, Xinggang and Wu, Yanmin and Zhao, Mingyuan and Cao, Zhenzhong and Zhang, Xiangkui and Ji, Xiangyang},
  journal={IEEE Transactions on Circuits and Systems for Video Technology},
  year={2025},
  publisher={IEEE}
}

@techreport{openai_gpt5_system_card_2025,
  title        = {GPT-5 System Card},
  author       = {{OpenAI}},
  institution  = {OpenAI},
  year         = {2025},
  month        = aug,
  type         = {Technical Report},
  url          = {https://cdn.openai.com/gpt-5-system-card.pdf},
  note         = {Accessed: 2025-11-14}
}

@inproceedings{delitzas2026funrec,
  title={{Reconstructing Functional 3D Scenes from Egocentric Interaction Videos}},
  author={Delitzas, Alexandros and Zhang, Chenyangguang and Gavryushin, Alexey and Di Mario, Tommaso and Sun, Boyang and Dabral, Rishabh and Guibas, Leonidas and Theobalt, Christian and Pollefeys, Marc and Engelmann, Francis and Barath, Daniel},
  booktitle=cvpr,
  year={2026}
}

@article{wei2023ov,
  title={Ov-parts: Towards open-vocabulary part segmentation},
  author={Wei, Meng and Yue, Xiaoyu and Zhang, Wenwei and Kong, Shu and Liu, Xihui and Pang, Jiangmiao},
  journal={Advances in Neural Information Processing Systems},
  volume={36},
  pages={70094--70114},
  year={2023}
}

@inproceedings{sun2023going,
  title={Going Denser with Open-Vocabulary Part Segmentation},
  author={Sun, Peize and Chen, Shoufa and Zhu, Chenchen and Xiao, Fanyi and Luo, Ping and Xie, Saining and Yan, Zhicheng},
  booktitle={Proceedings of the IEEE/CVF International Conference on Computer Vision},
  year={2023}
}
\end{document}